\documentclass[runningheads]{llncs}
\usepackage{csquotes}

\usepackage{graphicx}
\usepackage{listings}
\usepackage{algorithm}
\usepackage{algorithmic}
\newtheorem{define}{Definition}
\usepackage{amsmath,amssymb}
\usepackage{cleveref}
\usepackage{subcaption}
\usepackage{stmaryrd}
\usepackage{booktabs}

\usepackage{subcaption}
\begin{document}
\title{Is More Data All You Need? A Causal Exploration}
%
%
\author{Athanasios Vlontzos\inst{1} \and
Hadrien Reynaud\inst{1} \and
Bernhard Kainz\inst{1,2}}
%
\authorrunning{Vlontzos, Reynaud and Kainz}
%
\institute{Dept. of Computing, Imperial College London, London UK \and
 FAU Erlangen-Nuremberg, 
    Erlangen, Germany
}
%
\maketitle              
\begin{abstract}
Curating a large scale medical imaging dataset for machine learning applications is both time consuming and expensive. Balancing the workload between model development, data collection and annotations is difficult for machine learning practitioners, especially under time constraints. Causal analysis is often used in medicine and economics to gain insights about the effects of actions and policies. In this paper we explore the effect of dataset interventions on the output of image classification models. Through a causal approach we investigate the effects of the quantity and type of data we need to incorporate in a dataset to achieve better performance for specific subtasks. The main goal of this paper is to highlight the potential of causal analysis as a tool for resource optimization for developing medical imaging ML applications. We explore this concept with a synthetic dataset and an exemplary use-case for Diabetic Retinopathy image analysis. 

\end{abstract}
\section{Introduction}

Translating deep learning methods to new applications in the clinic often start with two questions that are very difficult to answer: How much data to collect and what data aspects need more attention than others to meet clinical performance expectations. In diagnostic settings, performance is often characterized by a biased metric, for example an expectation towards zero false negatives so that no signs of disease are missed but some leeway towards false positives, which can be mitigated with further diagnostic tests. However, this commonly leads to situations where  end-users request from machine learning practitioners to make specific interventions on well-performing  models, for example to make a deep neural network more sensitive towards one specific class of disease or to change predictions for a selected group of patients, while keeping its sensitivity and specificity intact for other classes. 

Active learning~\cite{budd2021survey}, where a probabilistic uncertainty based approach to decide on new samples to label, is one option to make such interventions on a working model but it has been shown that the introduced bias is not necessarily beneficial and might harm a model's specificity~\cite{farquhar2020statistical}. Furthermore, there is currently no method to estimate for how long further (active) learning should go ahead or how many more samples of specific classes have to be collected until the expected change can be observed. This has critical implications for practical translation of such methods into the clinical practice since time, costs and amount of data cannot be estimated in advance, which in turn conflicts with the need for data minimization as recommended by General Data Protection Regulations~\cite{voigt2017eu}.

We believe that methods like ours should be integral parts of regulation approval processes. As models undergo fine-tuning and retraining in the process of development and commercialization we should maintain the same high standards of accuracy and robustness. We need hence to be able to provide guarantees that the resulting ML models cannot degrade in performance. 



Therefore, the need arises to be able to reason about the data needs of an application and decide upon the best allocation of resources. Moving forward from the well known active learning paradigm we are looking at more targeted intervention scenario and provide in this paper a causal approach that allows to estimate how much \emph{extra} data is needed for targeted interventions on trained deep learning models. We show on a synthetic dataset and an exemplary large Diabetic Retinopathy (DR) medical imaging dataset how to use our approach.



\noindent\textbf{Contribution:} We treat the aforementioned scenarios as a counterfactual meta analysis upon a static model. Our goal is to highlight causal analysis as a potential alternative to active learning and showcase the powerful insights it could yield. Interestingly, we found that it is not always advantageous to increase the size of a dataset.

\noindent\textbf{Related Work:} 
Recent works on the field of model performance analysis have primarily been focused on determining the required number of samples to achieve.  \cite{figueroa2012predicting} developed a inverse power law model to predict model performance with different data sizes. \cite{cho2015much,beleites2013sample} performed empirical studies on the learning behavior of classifiers to determine sample size requirements.
None, however, of the above methods are able to determine the effect of interventions on individual samples, which is the primary concern of this investigation. 
Causality, on the other hand,  is the field of analysis of causal relationships between variables. 
The field was expanded to computer science by the works of, among others,  J. Pearl~\cite{Pearl2009}, however few works exist on the intersection of machine learning, medical imaging and causal analysis.
Recently, discussed as useful for medical image analysis~\cite{castro2020causality}, More commonly such approaches are found in fields like econometrics~\cite{imbens2015causal}, epidemiology~\cite{cuellar2020non} and clinical medicine\cite{louizos2017causal,richens2020improving,oberst2019counterfactual}. We are borrowing inspiration from these works to argue the potential advantages of causal analysis of medical imaging machine learning algorithms.

\section{Method}
\noindent\textbf{Preliminaries}
In this section we will first introduce key concepts and main mathematical tools of our analysis, finally in \Cref{results,discussion} we will detail our results and discuss the significance of our analysis. 

We work in the Structural Causal Models (SCM) framework. Chapter $7$ of \cite{Pearl2009} gives an in-depth discussion. For an up-to-date, self-contained review of  counterfactual inference and Pearl's Causal Hierarchy, see \cite{bareinboim20201on}.

\begin{define}[Structural Causal Model] \label{functional causal model}
\label{scmdef}
A structural causal model (SCM) specifies a set of latent variables $U=\{u_1,\dots,u_n\}$ distributed as $P(U)$, a set of observable variables $=\{v_1,\dots, v_m\}$, a directed acyclic graph (DAG), called the \emph{causal structure} of the model, whose nodes are the variables $U\cup V$, a collection of functions $F=\{f_1,\dots, f_n\}$, such that $v_i = f_i(PA_i, u_i), \text{ for } i=1,\dots, n,$ where $PA$ denotes the parent observed nodes of an observed variable.
\end{define}


The collection of functions $F$ and the distribution over latent variables induces a distribution over observable variables: $P(V=v) := \sum_{\{u_i \mid f_i(PA_i, u_i)\,=\,v_i\}}\\ P(u_i)$ 
In this manner, we can assign uncertainty over observable variables despite the fact the underlying dynamics are deterministic. 



Moreover, the $do$-operator forces variables to take certain values, regardless of the original causal mechanism. Graphically, $do(X=x)$ means deleting edges incoming to $X$ and setting $X=x$. Probabilities involving $do(x)$ are normal probabilities in submodel $M_x$: $P(Y=y \mid \text{do}(X=x)) = P_{M_x} (y)$.

\noindent\textbf{Counterfactual inference} \label{Section: two approaches to counterfactual inference}
The latent distribution $P(U)$ allows one to define probabilities of counterfactual queries, $P(Y_{y}=y) = \sum_{u \mid Y_{x}(u)=y} P(u).$ For $x \neq x'$ one can also define joint counterfactual probabilities,
$P(Y_{x}=y, Y_{x'}=y') = \sum_{u \mid Y_{x}(u)=y,\text{ }\& Y_{x'}(u)=y'} P(u).$
Moreover, one can define a counterfactual distribution given seemingly contradictory evidence. Given a set of observed evidence variables $E$, consider the probability $ P(Y_{{x}}={y}' \mid E = {e})$.

\noindent\textbf{Definition 2(Counterfactual):}\emph{The counterfactual sentence ``$Y$ would be $y$ (in situation $U=u$), had $X$ been $x$'', denoted $Y_{x}(u) = y$, corresponds to $Y=y$ in submodel $M_x$ for $U = u$.}

Despite the fact that this query may involve interventions that contradict the evidence, it is well-defined, as the intervention specifies a new submodel. Indeed, $ P(Y_{x}={y}' \mid E = {e})$ is given by~\cite{Pearl2009}
$ \sum_{{u}}
P(Y_{{x}}({u})={y}')P({u}|{e})\,.$
There are two main ways to resolve this type of questions; the Abduction-Action-Prediction paradigm and the Twin Network paradigm shown  respectively in ML literature among others in \cite{castro2020causality,vlontzos2021estimating}. In short given SCM $M$ with latent distribution $P(U)$ and evidence ${e}$, the conditional probability $P(Y_{{x}} \mid {e})$ is evaluated as follows: 1) \emph{Abduction:} Infer the posterior of the latent variables with evidence ${e}$ to obtain $P(U \mid {e})$, 2) \emph{Action:} Apply $\text{do}({x})$ to obtain submodel $M_{x}$, 3) \emph{Prediction:} Compute the probability of $Y$ in the submodel $M_{x}$ with $P(U \mid {e})$. Meanwhile the Twin Network paradigm casts the resolution of counterfactual queries to Bayesian feed-forward inference by extending the SCM to represent both factual and counterfactual worlds at once~\cite{balke1994counterfactual}.

\begin{figure}[htb]
    \centering
 {
        \includegraphics[width=0.5\linewidth]{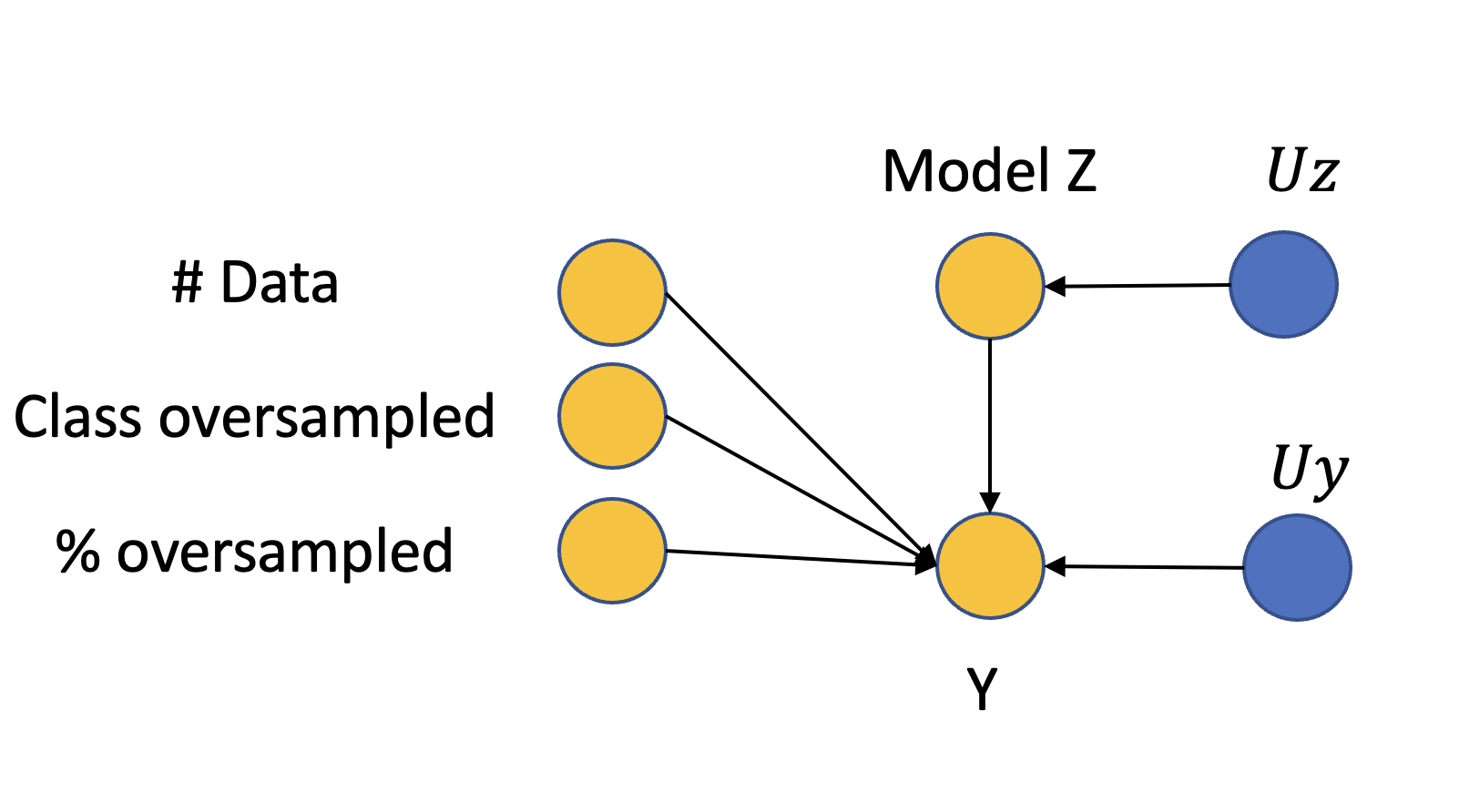}
        }
  
    \caption{A DAG showing the causal relationships between the factors we are analyzing - Yellow: Observed variables - Blue: Unobserved Variables. For simplicity we intervene on one of the possible treatment variables during each experiment. }
    \label{figure: sample dag}
\end{figure}

\noindent\textbf{Analysis framework\label{methods}}
As discussed above and expanded in \Cref{discussion} there exists a wide range of possible methods that enable accurate estimation of counterfactual quantities in real world scenarios~\cite{schwab2018perfect,castro2020causality,vlontzos2021estimating}. For the scope of this paper we are assuming perfect knowledge and as such we revert to the mathematical tools that counterfactual inference methods approximate. We are going to be mainly concerned about the effect of datasets on the classification outcome of the samples. We treat the model architecture and its hyper parameters as confounders that affect only the outcome and are invariant between different treatments of our dataset. We intervene on the size and composition of our dataset leaving all other parameters the same. In \Cref{figure: sample dag} we show a sample directed acyclic graph (DAG) for the causal relationships between the variables we take into account for our analysis. For simplicity for each counterfactual we explore only interventions on one of the possible treatments.

 Our main research question is a counterfactual one; given a sample was incorrectly classified would this sample be correctly classified if we had trained our model with a different dataset? Mathematically this resembles the probability Sufficiency  $P_S = P(Y_{X=T}=y \mid X=T', Y=y')$ and  can be described as: 

\begin{equation}
\label{p_s_eq}
    P(Y=1, do(X=\hat{\mathcal{D}}) \mid Y=0, X=\mathcal{D},Z)
\end{equation}
where Y is the outcome of whether or not the sample was correctly classified; X represents the treated dataset $\mathcal{D}$ and Z the confounders like the architecture of the model. 
Our analysis regards the model as a black box and its underlying architecture complexity is not a constraint. We use this question as an example for a potential causal analysis on a medical image model. We highlight that multiple counterfactual questions can be asked, concerning for example the model architecture or a specific feature of the data. We chose interventions on the overall statistics of the dataset as an entry level counterfactual question with evident real world impact on ML practice.

\section{Evaluation\label{evaluation}}

\noindent\textbf{Datasets:}
We use two datasets, one synthetic and one real medical imaging database. As synthetic data we use the MorphoMNIST~\cite{castro2019morphomnist} dataset. This dataset provides a series of morphological operations upon the digits of the well known MNIST dataset~\cite{lecun1998gradient}. Each of the generated datasets is  modified by random fractures and swellings. In order to control this perturbation in our dataset throughout the experiments each sample retains the same morphological perturbation. As this dataset is fully controlled and synthetic we can draw new never seen before samples each time, simulating new data gathering regimes.

Furthermore, we chose  Kaggle's Retinopathy open source dataset~\cite{diabeticretinopathydetection} as medical imaging dataset. We treat each of the images as a gray scale image and resize it down to $128\times128 px$. As this is a multi-class dataset with heavy imbalance, we focus on the following labels for Diabetic Retinopathy (DR) levels : "No DR", "Mild" and "Moderate" cases.

\noindent\textbf{Model architecture:}
As a first step we train a classifier with the full training set. The goal of this step is to determine an architecture and set of hyper-parameters that are able to provide acceptable classification results. Having determined such parameters we fix them for each counterfactual query as to determine the true causal effect of our dataset interventions on the probability of correct classification of a sample. For the synthetic case we opt for a 3 layered multi-layered Perceptron while for the medical use case a simple multi scale residual convolutional net. The residual network is comprised of 4 residual blocks with decreasing resolution.  While in the synthetic case we opt for a simple model and in the medical one with a significantly more complex one, we note that this is not a parameter that imposes any constraints on our analysis. Exact architectures can be found in the Appendix and implementations will be made public together with the codebase.

\noindent\textbf{Interventions:}
Interventions are focused on the size and composition of each dataset. First we explore the effect of the size of the dataset. Given that both the synthetic and medical tasks have an abundance of data we create a series of datasets with $[100,1000,5000,10000]$ samples for the synthetic dataset and $[100,500,1000,2000,4000]$ samples from the medical dataset. We use the intervened datasets to train our models and a static test dataset that includes $20\%$ of the full dataset. This serves as a gold standard evaluation set. Moreover, we intervene upon the dataset by modulating the number of samples in selected classes as well as the percentage of the base dataset by which we will increase a class. In other words given a base dataset of length $N_{dataset}$ with approximate class balance and an upsampling percentage of $10\%$ with class $2$, we will add $10\% \times N_{dataset}$ that has unseen samples of class $2$. We explore upsampling all available classes by one of the following percentages $0\%, 5\%, 10\%, 20\%, 30\%, 50\%$

For example assuming an intervention of dataset size from $100$ to $1000$ we are going to be calculating the measure: 
\begin{equation}
    \begin{split}
Query  & = P(Y_{do(N=1000)}=1 \mid Y=0, N=100, Z) \\
 & = \frac{P(Y=1,N=1000,Z) \wedge P(Y=0,N=100,Z)}{P(Y=0,N=100,Z)}
\end{split}
\end{equation}
Note that this simplification of the query is possible as we assume full knowledge of the simple DAG of \Cref{figure: sample dag}. In more complex scenarios proper controlling for front- and back-door criteria should be happen. In addition in our simple scenario each of the aforementioned probabilities can be calculated as the average number of events \emph{i.e.} for a model trained in the regime of confounders $Z$ and $1000$ training samples:
\begin{equation}
    P(Y=1,N=1000,Z) = \frac{\# \text{correctly classified samples}}{\# \text{total samples}}
\end{equation}

\section{Results\label{results}}

\begin{figure}
    \centering
   \begin{minipage}{.4\linewidth}
      \centering
      \includegraphics[width=\linewidth]{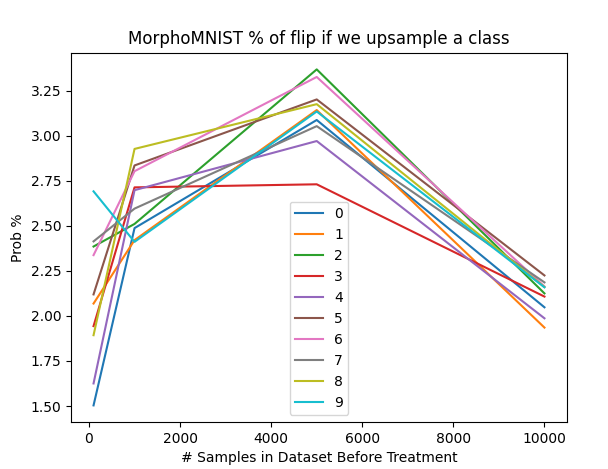}
          \caption{Probability of flip; Treatment: Upsampling class - see legend\label{fig:morphomnist_upsampleclass}}
    \end{minipage}%
    \quad
    \begin{minipage}{.4\linewidth}
      \centering
        \includegraphics[width=\linewidth]{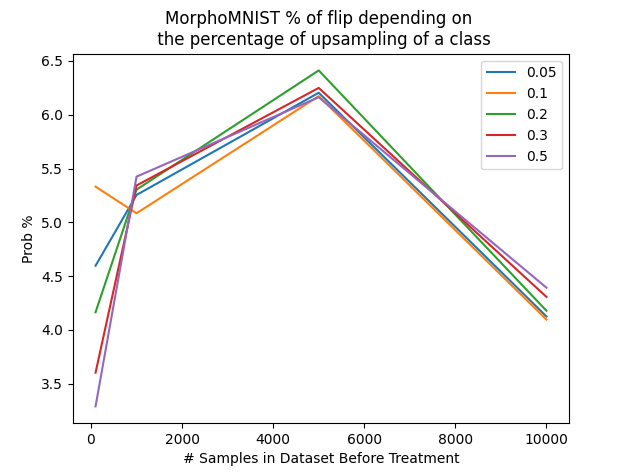}
                \caption{Probability of flip; Treatment: Upsampling percentage of base dataset - see legend \label{fig:morphomnist_upsamplepercentage}}
    \end{minipage} 

\end{figure}

\noindent\textbf{Synthetic Data} First we evaluate the synthetic task.
 In \Cref{fig:random class more data} we show the effect of including more data regardless of the type of data given a base dataset. We notice for example that if we had 5k samples and switched to a dataset with 10k incorporating randomly selected samples that would give us only a 9\% chance of correctly classifying a sample that was previously misclassified. As a general observation we can see that the more samples are contained in our \emph{base dataset} the smaller the probability of a specific sample being correctly classified after the intervention of including more images. A proxy to this insight can be seen in \Cref{tab:f1 morphomnist} where we show the average F1 score for all classes for different sized datasets. We see that indeed the overall improvement between 5 and 10 thousand samples is quite small. Meanwhile in \Cref{fig:morphomnist_upsampleclass,fig:morphomnist_upsamplepercentage} we show the probability of a sample ``flipping" from being misclassified to being correctly classified versus the class we upsample and the percentage of upsampling. In both cases we look at any given sample without taking into account the ground truth label of it. It is evident from both figures that there is no single class or percentage of upsampling that is key for this dataset -- in other words, there is no category of samples that contain key information to help the classification task, rather all classes seem to be necessary. 
 
 \begin{table}[htb]
\caption{MorphoMNIST if we change the number of samples regardless of class and percentage. (A) The probability of changing a misclassified sample to a correctly classified one is shown.(B) F1 performance of  different sized base datasets}
\begin{subtable}{.5\linewidth}

      \caption{\label{fig:random class more data} }
      \centering
        
\begin{tabular}{@{}lllll@{}}
\toprule
\multicolumn{1}{c}{\begin{tabular}[c]{@{}c@{}}\# of samples\\ From / To\end{tabular}} & 100    & 1000    & 5000    & 10000   \\ \midrule
100                                                                                   & 0      & 23.17\% & 27.18\% & 27.72\% \\
1000                                                                                  & 4.17\% & 0       & 17.57\% & 18.93\% \\
5000                                                                                  & 3.95\% & 7.22\%  & 0       & 9.46\%  \\
10000                                                                                 & 3.94\% & 7.20\%  & 7.25\%  & 0       \\ \bottomrule
\end{tabular}
    \end{subtable}%
    \quad
\begin{subtable}{.5\linewidth}
      \centering
        \caption{\label{tab:f1 morphomnist}}
        \begin{tabular}{@{}ll@{}}
\toprule
\multicolumn{1}{c}{} & F1     \\ \midrule
100                  & 30.606\% \\
1000                 & 75.97\% \\
5000                 & 85.54\% \\
10000                & 86.84\%  \\ \bottomrule
\end{tabular}

    \end{subtable} 

\end{table}

 Thus far we have increased class samples and adding more data to our base dataset without taking into consideration the true class of the misclassified data. In the next step we look into informed interventions where we include a larger number of samples where the majority of them are from the class which was misclassified.
 In \Cref{tab:informed_interv_mnist} we show the effects of different dataset sizes and percentages of upsampling on the probability of a specific sample being correctly classified after the intervention. Compared to incorporating more data randomly or upsampling a class regardless of the misclassfied class we observe a significantly increased effect. If, conversely, we do not specify the extend we upsample the misclassified class we average $67.56\%$ probability of correctly classifying a sample after the intervention across all possible percentages and dataset sizes. It is evident, thus, that we can obtain a higher probability of sufficiency if our interventions on the dataset are targeted. We further note that during this analysis we look at the inverse probability of incorrectly classifying a sample that was correctly determined before treatment, in all our cases this probability did not exceed $3-4\%$ indicating that our chosen interventions did not have a negative effect upon the model performance.
 
 \begin{table}[]
 \begin{subtable}{.5\linewidth}

\caption{}
 \centering
\begin{tabular}{@{}lllll@{}}
\toprule
\multicolumn{1}{c}{} & 100   & 1000  & 5000  & 10000 \\ \midrule
full dataset         & 7.71  & 21.18 & 23.35 & 24.48 \\
100                  & 0     & 29.16 & 30.12 & 30.32 \\
1000                 & 19.34 & 0     & 24.30 & 24.92 \\
5000                 & 18.70 & 17.41 & 0     & 18.58 \\
10000                & 18.49 & 16.94 & 16.69 & 0     \\ \bottomrule
\end{tabular}
\end{subtable}
\quad
\begin{subtable}{.5\linewidth}
      \caption{}
      \centering 
      \begin{tabular}{@{}llllll@{}}
\toprule
\multicolumn{1}{c}{} & 0.05  & 0.1   & 0.2   & 0.3   & 0.5   \\ \midrule
full dataset         & 8.13  & 9.21  & 11.49 & 13.86 & 15.72 \\
0.05                 & 0     & 11.50 & 14.36 & 17.16 & 19.61 \\
0.1                  & 9.99  & 0     & 14.64 & 17.21 & 19.66 \\
0.2                  & 10.64 & 12.03 & 0     & 17.53 & 19.83 \\
0.3                  & 11.13 & 12.68 & 15.27 & 0     & 20.02 \\
0.5                  & 11.86 & 13.27 & 15.87 & 18.27 & 0     \\ \bottomrule
\end{tabular}
\end{subtable}
\caption{Informed Interventions (A) the effect of different dataset sizes (B) the effect of different upsampling percentages \label{tab:informed_interv_mnist} }

\end{table}

\noindent\textbf{Retinopathy:} we follow a similar analysis for the medical image data where we classify some of the most abundant categories of the open source Retinopathy dataset.  In the interest of space we only include some results here, full tables can be found in the Appendix. 
In \Cref{tab:informed_interv_retina} we show the two most interesting results. In this real world dataset we observe that incorporating more of the \textit{moderate DR} class leads to all together better classification performance regardless of the dataset size. On the other hand modulating the overall number of samples under an informed sampling regime seems to be driven primarily by our informed sampling than the actual changes in number of data. We note that increasing the datasize from 100 to 2000 by randomly sampling from the available classes only provides a $\sim 10\%$ chance of a sample flipping. Medical imaging datasets can offer interesting insights if looked under a causal prism. It is possible to identify inter-dependencies of classes and features and hence able to plan the dataset acquisition and annotations more efficiently.

\section{Discussion\label{discussion}}
We analyzed the effect of dataset size and composition on the probability of a specific misclassified sample to become correctly classified after our interventions. We have observed a wide expected range for this causal probability. If used in practice to analyse a phenomenon and determine the best allocation of resources certain thresholds that make sense  have to be determined by the users. Contrary to the well known active learning paradigm where we are interested in the effect of an intervention on the overall metrics of our task; by assuming a causal perspective we are able to estimate the effect of interventions or counterfactuals on an individual sample. This ability, enables a finer grain analysis of our interventions and their effects. 
For the purposes of our analysis we have assumed complete knowledge of the behavior of our models under different data regimes. This however, is not a valid assumption in real life model development. In such cases, the practitioner could employ a method from literature to estimate or bound the above probability of causation. 

If we do not condition on the knowledge that the sample was initially misclassified and we are solely interested in the interventional probability if it will be correctly classified, we aim to learn the conditional average treatment effects: $\mathbb{E}(Y_{X=1} \mid Z) - \mathbb{E}(Y_{X=0} \mid Z)$. 
Examples of methods that can estimate these include PerfectMatch \cite{schwab2018perfect}, DragonNet \cite{shi2019adapting}, PropensityDropout \cite{alaa2017deep}, Treatment-agnostic representation networks (TARNET) \cite{shalit2016estimating}, Balancing Neural Networks \cite{johansson2016learning}. Other machine learning approaches to estimating interventional queries made use of GANs, such as GANITE \cite{yoon2018ganite} and CausalGAN \cite{kocaoglu2018causalgan}, Gaussian Processes  \cite{witty2020causal,alaa2017bayesian}, Variational Autoencoders  \cite{louizos2017causal}, and representation learning \cite{zhang2020learning,assaad2021counterfactual,yao2018representation}.

If, on the other hand, we wish to answer the same query  in \Cref{p_s_eq} we need to utilize methods able to handle counterfactual queries.  Recent work \cite{pawlowski2020deep} proposes normalising flows and variational inference to compute counterfacual queries using abduction-action-prediction. 
 \cite{oberst2019counterfactual} used the Gumbel-Max trick to estimate counterfactuals, again using abduction-action-prediction. While this methodology satisfies generalisations of identifiability constraints. Additional work by~\cite{cuellar2020non} devised a non-parametric method to compute the Probability of Necessity using an influence-function-based estimator. A limitation of this approach is that a separate estimator must be derived and trained for each counterfactual query. Finally, \cite{vlontzos2021estimating} estimate counterfactual probabilities via means of a deep twin network while imposing identifiability constraints in the case of binary treatments and outcomes.

A major difference between the aforementioned causally enabled methods to prior active learning is the ability of causal methods to answer individual treatment effect queries. Utilizing methods like \cite{vlontzos2021estimating,shalit2017estimating,pearl21individual} one is able to estimate the effect of a treatment at an individual level. In our example case that translates the assessment of the probability of a sample to be correctly identified after the treatment in the model's training regime is applied. This increase in analytical resolution admits reasoning about the allocation of resources at a level that was not possible with aggregate level statistics.

 \begin{table}[]
 \begin{subtable}{.5\linewidth}

\caption{}
 \centering
\begin{tabular}{@{}llll@{}}
\toprule
\multicolumn{1}{c}{} & Healthy   & Mild     & Moderate   \\ \midrule
Healthy       & 0  & 12.03 & 18.75  \\
Mild        &  12.95    & 0 & 14.74  \\
Moderate               & 16.52& 11.31  & 0     \\ \bottomrule
\end{tabular}
\end{subtable}
\quad
\begin{subtable}{.3\linewidth}
      \caption{}
      \centering 
      \begin{tabular}{@{}lllll@{}}
\toprule
\multicolumn{1}{c}{} & 100  & 500   & 1000   & 2000      \\ \midrule
100                & 0     & 18.28 & 18.16 & 19.77  \\
500                  & 18.23  & 0     & 18.00 & 19.70  \\
1000                  &18.33 & 18.04 & 0     & 18.79  \\
2000                 & 18.88 & 18.71 & 17.59 & 0      \\ \bottomrule
\end{tabular}
\end{subtable}
\caption{Dataset Interventions on the Retina dataset (A) the effect of different upsampling classes  (B) the effect informed interventions and dataset sizes   \label{tab:informed_interv_retina} }
\end{table}

\section{Conclusion\label{conclusion} \& Future Work}
With this paper we hope to stimulate a new topic of discussion in the ML and Medical imaging community around causal analysis and how it can help us optimize  resource allocations. Being able to quantify the per sample effect of an intervention is necessary to better understand a given task.
Besides economical constraints, the proven environmental impact of our field~\cite{dhar_2020} means that we  cannot opt for increasingly larger models when the expected returns are minimal. Causally analyzing the task at hand can provide estimates of performance vs. computational and economical resources without the need to run the experiments.

This work can be considered as an introductory preparatory work. While other methods provide argumentation in favor of using causality to enable robust and explainable ML algorithms we showcase the use of causality and causal analysis in a business intelligence task. We believe that future work can include methodological contributions as causally enabled methods to provide estimates of the amount and distributions of data in a dataset. Moreover, we aim at working together with regulators to set out the appropriate thresholds of confidence such that our analysis can inform regulatory procedures, ensuring that a given model is not going to be degraded upon retraining or fine-tuning with data that abide by the identified causal relationships.

\section{Acknowledgments}
The authors would like to acknowledge and thank
 the  MAVEHA (EP/S013687/1) project,  Ultromics Ltd. and the UKRI Centre for Doctoral Training in Artificial Intelligence for Healthcare (EP/S023283/1). The authors also received GPU donations from NVIDIA.

\bibliographystyle{splncs04}
\bibliography{bibliography.bib}

\end{document}